\documentclass[conference]{IEEEtran}
\IEEEoverridecommandlockouts
\usepackage{booktabs}
\usepackage{multirow}
\usepackage[table]{xcolor}   
\usepackage{cite}
\usepackage{amsmath,amssymb,amsfonts}
\usepackage{algorithmic}
\usepackage{graphicx}
\usepackage{textcomp}
\usepackage{xcolor}
\usepackage{multirow}
\usepackage{subfig}
\def\BibTeX{{\rm B\kern-.05em{\sc i\kern-.025em b}\kern-.08em
    T\kern-.1667em\lower.7ex\hbox{E}\kern-.125emX}}

\ifodd 0
\newcommand{\rev}[1]{{\color{green}#1}} 
\newcommand{\needrev}[1]{{\color{red}#1}} 

\else
\newcommand{\rev}[1]{#1}
\newcommand{\needrev}[1]{#1}

\fi

\begin{document}

\newcommand{\name}{FC-MoE }

\title{Conflict-Aware Federated Fine-Tuning of Large Language Models with Mixture-of-Experts
}





    




\author{\IEEEauthorblockN{\hspace{1em}Yijun Lu\IEEEauthorrefmark{2}}
\IEEEauthorblockA{\textit{\hspace{1em}Waseda University, Tokyo, Japan} \\
\hspace{1em}yijun@ruri.waseda.jp}
\and
\IEEEauthorblockN{\hspace{2em}Zihan Fang\IEEEauthorrefmark{2}}
\IEEEauthorblockA{\hspace{2em}\textit{City University of Hong Kong, China} \\\hspace{2em}zihanfang3-c@my.cityu.edu.hk}
\and
\IEEEauthorblockN{\hspace{2em}Pengpeng Qiao}
\IEEEauthorblockA{\hspace{2em}\textit{Institute of Science Tokyo, Tokyo, Japan} \\
\hspace{2em}peng2qiao@gmail.com}
\and
\IEEEauthorblockN{Zheng Lin\IEEEauthorrefmark{1}}
\IEEEauthorblockA{\textit{The University of Hong Kong, China} \\
linzheng@eee.hku.hk}
\and
\IEEEauthorblockN{\hspace{-3em}Jing Yang}
\IEEEauthorblockA{\textit{\hspace{-3em}Universiti Malaya, Malaysia} \\
\hspace{-3em}s2147529@siswa.um.edu.my}
\and
\IEEEauthorblockN{Yuxin Zhang}
\IEEEauthorblockA{\textit{Fudan University, Shanghai, China} \\
yxzhang24@m.fudan.edu.cn}
\and
\IEEEauthorblockN{\hspace{2em}Por Lip Yee}
\IEEEauthorblockA{\hspace{2em}\textit{Universiti Malaya, Malaysia} \\
\hspace{2em}porlip@um.edu.my}
\and
\IEEEauthorblockN{\hspace{3em}Zhe Chen}
\IEEEauthorblockA{\hspace{3em}\textit{Fudan University, Shanghai, China} \\
\hspace{3em}zhechen@fudan.edu.cn}
\and
\IEEEauthorblockN{\hspace{4em}Jun Luo}
\IEEEauthorblockA{\textit{\hspace{4em}Nanyang Technological University} \\
\hspace{4em}junluo@ntu.edu.sg}

\thanks{\IEEEauthorrefmark{2} Yijun Lu and Zihan Fang contributed equally to this work.}
\thanks{\IEEEauthorrefmark{1} denotes the corresponding author.}
}

\maketitle

\begin{abstract}
The continuous scaling of large language models (LLMs) incurs prohibitive computational costs,  making Mixture-of-Experts (MoE) a scalable alternative for efficient fine-tuning via sparse activation. While federated learning (FL)  emerges as the paradigm for privacy-preserving collaborative optimization, integrating MoE into FL under data heterogeneity may trigger conflicting expert optimizations. Client-specific data distributions forces same-indexed experts to optimize under inconsistent or even conflicting feature-label correlations. This mismatch induces destructive interference during aggregation, thus destabilizing the optimization trajectory and degrading model performance.
To address this issue, we propose FC-MoE, a federated conflict-aware framework for MoE fine-tuning. It employs an importance-aware weighting scheme to prioritize reliable local update and utilizes gradient consensus projection to suppress conflicting updates, ensuring a stable global optimization path. Moreover, a local knowledge retention mechanism further preserves specialized client expertise by re-anchoring domain-specific residuals. Extensive experiments demonstrate that \name  accelerates convergence and  enhances both global and local model performance in non-IID federated environments.
\end{abstract}


\section{Introduction}

Large language models (LLMs), such as GPT~\cite{achiam2023gpt}, LLaMA~\cite{touvron2023llama}, and DeepSeek~\cite{liu2024deepseek}, have achieved transformative success across academia and industry by scaling model parameters and training data, which enables the models to capture intricate linguistic patterns during pre-training and effectively generalize to diverse downstream tasks through fine-tuning~\cite{ chen2025empower,lin2025hsplitlora,fang2024automated,zhang2026hera}.
However, continuous scaling of dense transformer models is increasingly hindered by the prohibitive computational cost required for large-scale pre-training and downstream adaptation~\cite{he2024towards, du2024sida, lin2025leo,wang2024scaling}. 
To address this limitation, Mixture-of-Experts (MoE) architectures~\cite{lin2026moe, dai2024deepseekmoe, fedus2022switch,fang2026hfedmoe} have emerged as a scalable alternative. MoE employs a gating network to selectively route each token to a sparse top-k subset of specialized experts while leaving the remaining experts inactive. This sparse activation mechanism enables MoE models to retain most of their representational capacity while achieving performance comparable to large dense counterparts with significantly reduced computational overhead~\cite{dai2024deepseekmoe, fedus2022switch, mei2024fedmoe, guo2021pfl,fang2026aggregation}, making MoE-based LLM a promising foundation for efficient fine-tuning~\cite{liu2025cmos}.


The real-world deployment of MoE-based LLMs demands massive volumes of data, which frequently contain sensitive details such as personal medical records and financial information~\cite{ge2025optimized, khan2024asmf}. Because strict privacy regulations prohibit sharing these private datasets across clients, traditional centralized training paradigms face a critical bottleneck. Federated Learning (FL) has therefore emerged as a promising paradigm that enables collaborative model optimization without compromising data privacy~\cite{latif2025mitigating,lin2024fedsn,hu2024accelerating,zhang2024satfed}. Combining MoE’s sparse activation with the decentralized framework of FL~\cite{mei2024fedmoe}, federated MoE-based LLM fine-tuning allows each client to locally activate and fine-tune only the sparse subset of experts relevant to its private datasets, and the updated experts are then aggregated on the server to construct an improved global MoE model, facilitating the integration of diverse knowledge across diverse clients.

\begin{figure*}[tb]
\centering
\includegraphics[width=14cm]{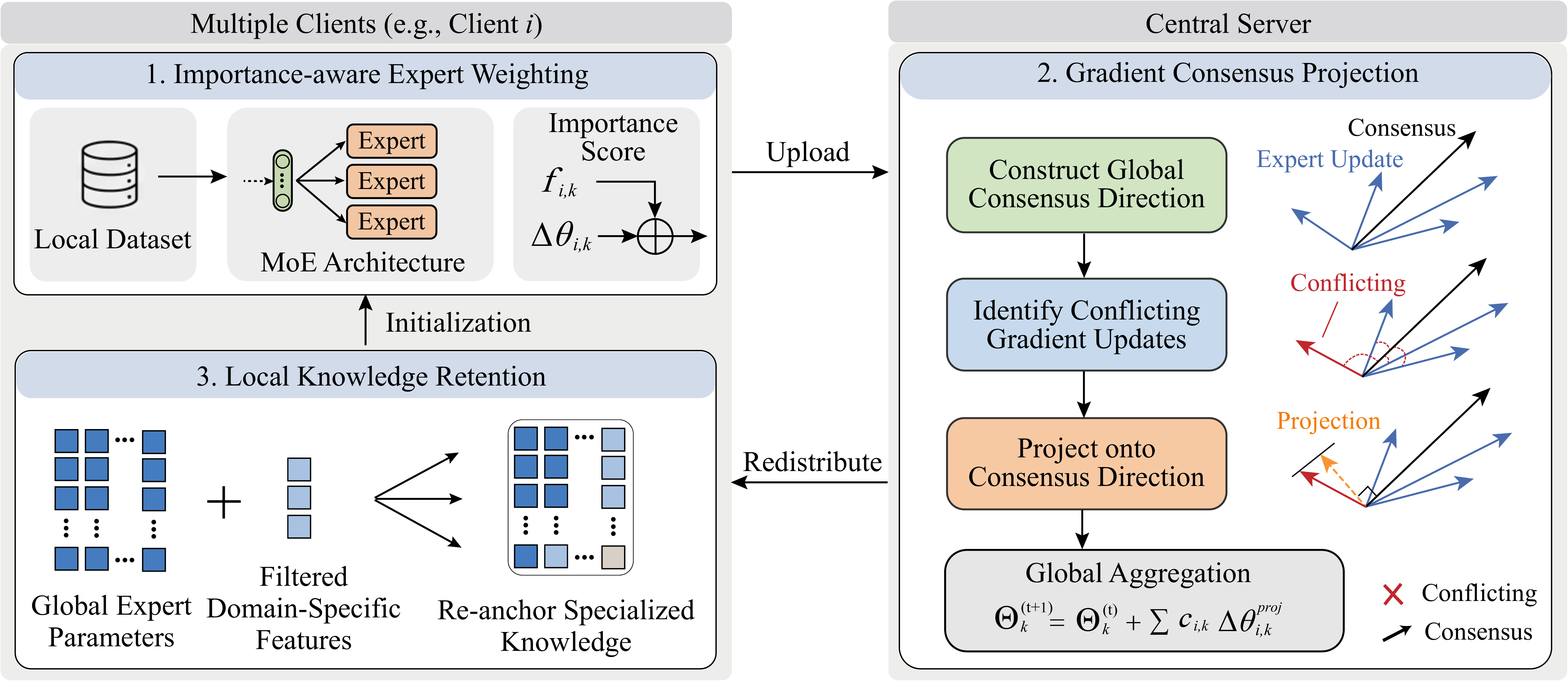}
\caption{The overview of \name framework.}
\label{fig:framwork}
\end{figure*}

While the integration of MoE into FL frameworks holds significant promise for efficient LLM fine-tuning, \rev{it encounters critical aggregation challenges due to inherent data heterogeneity}, where clients typically possess non-identically distributed (non-IID) data stemming from variations in local domains or task-specific requirements~\cite{zhu2021federated, kairouz2021advances, xie2025dflmoe}.
Given that the MoE routing mechanism dynamically directs tokens to experts based on client-specific data distributions, this heterogeneity causes same-indexed experts to be optimized under inconsistent or even conflicting feature–label correlations, leading to disparate expert activation and optimization patterns across clients. 
Consequently, updates for the same-indexed expert from various clients often become misaligned or even contradictory~\cite{feng2025pm}, inducing destructive interference during aggregation as existing solutions~\cite{mei2024fedmoe, guo2021pfl, xie2025dflmoe} typically assume all expert updates are mutually compatible.
This interference destabilizes the optimization trajectory, degrading the convergence stability and the performance of the global MoE model. 

To tackle expert conflicts induced by non-IID data distributions, we propose a \underline{f}ederated \underline{c}onflict-aware framework for \underline{MoE}-based LLM fine-tuning, named FC-MoE.
The framework aligns global consensus with local expert specialization by identifying conflicting updates and projecting their gradients toward a consensus direction during aggregation using an importance-aware weighting scheme, ensuring a stable and consistent optimization trajectory for the global MoE model.
To prevent the loss of client-specific knowledge, we further introduce a local knowledge retention strategy that allows each client to re-anchor the specialized knowledge during local initialization, effectively mitigates expert conflicts without sacrificing local model performance.

\section{Method}
\subsection{Overall Architecture of \name}
As shown in Fig.~\ref{fig:framwork}, the proposed \name comprises three key modules: importance-aware expert weighting, gradient consensus projection, and local knowledge retention.
Each client first computes importance scores for its active experts through importance-aware expert weighting and transmits these scores to the server along with local parameter updates.
The server then constructs a global consensus direction based on the received expert weights and projects conflicting expert gradients onto this direction to suppress destructive interference.
Finally, the resulting global consensus is redistributed to each client, where local knowledge retention is applied to re-incorporate the filtered domain-specific features into the local model initialization for the next training round.

\subsection{Importance-aware Expert Weighting}
Existing aggregation schemes typically assume that all local updates are mutually compatible and weight them solely based on client dataset size. Such indiscriminate averaging can induce destructive interference as misaligned or contradictory local experts across clients are allowed to negatively impact the global model. To mitigate this interference and guide a stable global aggregation, we introduce a dynamic weighting scheme that explicitly quantifies the importance of each local expert update.

In the MoE architecture, an expert's activation frequency reflects its statistical reliability with respect to the client's specific data distribution, as frequently activated experts tend to produce more representative gradients.
Activation frequency captures usage patterns but fails to reflect the extent of knowledge acquisition. Therefore, we further incorporate the gradient magnitude (i.e., the norm of the parameter update) as a measure of how much the expert has adapted to the local data, providing a direct geometric characterization of its learning intensity.
Specifically, after each local training, client $i$ calculates the parameter update for each active expert $k$ as $\Delta \theta_{i,k} = \theta_{i,k}^{local} - \theta_{k}^{global}$, where $\theta_{i,k}^{local}$ and $\theta_{k}^{global}$ are the local and global expert parameters. The importance score $c_{i,k}$ for expert $k$ is then defined as
\begin{equation}
c_{i,k} = f_{i,k} \cdot \|\Delta \theta_{i,k}\|,
\end{equation}
where $f_{i,k}$ represents the total activation count of expert $k$ under the local gating network on client $i$'s local dataset $\mathcal{D}_i$, prioritizing experts that are heavily utilized by local data. The term $\|\Delta \theta_{i,k}\|$ is the norm of the expert's parameter update, capturing the intensity of local adaptation.
By calculating the product of these two factors, this importance-aware expert weight ensures that the global aggregation is driven by experts that are both highly relevant and significantly optimized, \rev{providing a reliable indicator to mitigate expert conflicts during global aggregation.}
Client $i$ then uploads the tuple $(\Delta \theta_{i,k}, c_{i,k})$ to the central server.

\subsection{Gradient Consensus Projection}
Standard federated aggregation schemes (e.g., FedAvg) perform direct averaging of local updates, which results in destructive interference for MoE-based LLM under non-IID data distributions. To address this, we shift the aggregation from blind averaging to consensus-guided geometric projection. 
\rev{As gradient updates of local parameters mathematically represent the optimization trajectories of client-specific objectives, they can be used as a geometric indicator to detect conflicting expert optimizations.} By establishing a global consensus trajectory as a reference anchor, the server can explicitly quantify the geometric compatibility of each local update, encouraging the model to converge toward a generalized global optimum rather than collapsing under conflicting local biases.

Specifically, the server calculates a global consensus direction $\Delta \bar{\theta}_k$ for each expert $k$, which represents the dominant optimization direction driven by the majority of the relevant global data distribution. This consensus is calculated by the importance-weighted aggregation of local updates:
\begin{equation}
\Delta \bar{\theta}_k = \sum_{i=1}^N \frac{c_{i,k}}{\sum_{j=1}^N c_{j,k}} \Delta \theta_{i,k},
\end{equation}
where $\mathcal{S}_k$ denotes the set of clients that activated expert $k$ in the current round.

\needrev{Based on this reference anchor}, the server quantifies the geometric interference between each local update $\Delta \theta_{i,k}$ and the global consensus $\Delta \bar{\theta}_k$, where a conflict is defined by a negative cosine similarity $\text{cos}(\Delta \theta_{i,k}, \Delta \bar{\theta}_k) < 0$.
To mitigate expert conflicts without completely discarding the client's local learning, we project the conflicting local update $\Delta \theta_{i,k}$ onto the normal plane of the consensus direction $\Delta \bar{\theta}_k$ as
\begin{equation}
\Delta \theta_{i,k}^{proj} = \begin{cases} \Delta \theta_{i,k} - \frac{\Delta \theta_{i,k} \cdot \Delta \bar{\theta}_k}{\|\Delta \bar{\theta}_k\|^2} \Delta \bar{\theta}_k, & \text{if  } \text{cos}(\Delta \theta_{i,k}, \Delta \bar{\theta}_k) < 0 \\ \Delta \theta_{i,k}, & \text{otherwise} \end{cases}
\end{equation}

Finally, the server aggregates the projected updates using the normalized contribution scores to update the global expert parameters, ensuring that the global MoE model follows a stable and consistently improving optimization trajectory:
\begin{equation}
\Theta_k^{(t+1)} = \Theta_k^{(t)} + \sum_{i=1}^N \frac{c_{i,k}}{\sum_{j=1}^N c_{j,k}} \Delta \theta_{i,k}^{proj}.
\end{equation}

\begin{figure}[t]
\centering
\includegraphics[width=9cm]{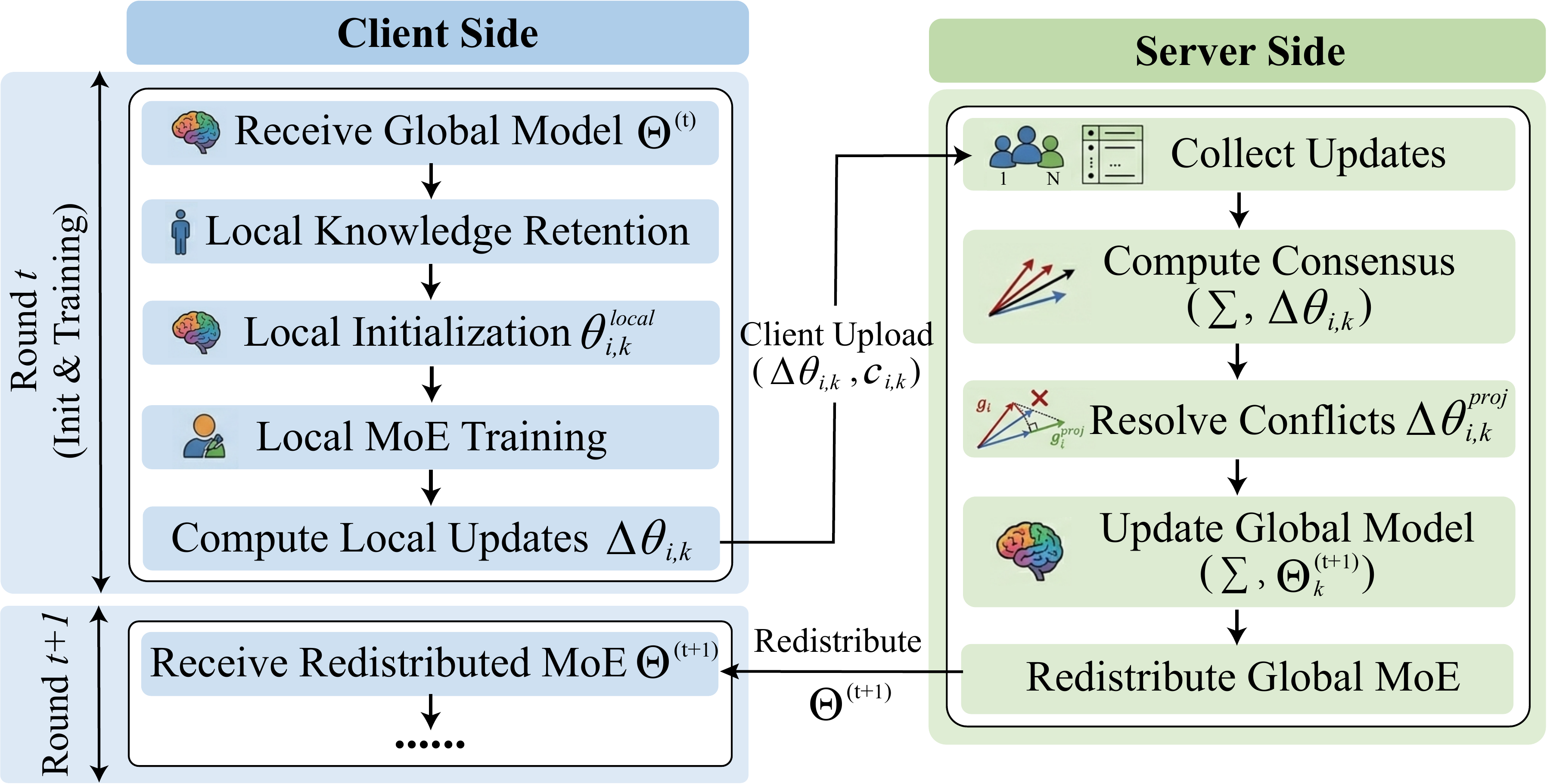}
\caption{The overall workflow of conflict-aware federated MoE-based LLM fine-tuning.}
\label{fig:workflow}
\vspace{-1ex}
\end{figure}

\subsection{Local Knowledge Retention}
While the gradient consensus projection facilitates global convergence by mitigating conflicting components, it may inevitably strip away certain domain-specific features required for optimal local performance. To prevent the degradation of local expert specialization, \name enables each client to isolate its specialized gradient residuals locally and strategically re-incorporate them into its local optimization in the next training round. The overall workflow is illustrated in Fig.~\ref{fig:workflow}.

Specifically, upon receiving the aggregated global model $\Theta^{(t+1)}$ from the server at the beginning of round $t+1$, each client locally derives a filtered residual for each expert $k$. To extract its personalized domain knowledge, the client projects its previous local update $\Delta \theta_{i,k}$ onto global expert parameters:
\begin{equation}
R_{i,k} = \frac{\Delta \theta_{i,k} \cdot \Theta_k^{(t+1)}}{\|\Theta_k^{(t+1)}\|^2} \Theta_k^{(t+1)}
\end{equation}

Before starting the next round local training, client $i$ explicitly anchors this localized knowledge by incorporating a scaled version of the residual back into its local expert parameters:
\begin{equation}
\theta_{i,k}^{local} = \Theta_k^{(t+1)} + \lambda R_{i,k}
\end{equation}
where $\lambda$ serves as a retention coefficient that controls the intensity of local specialization. 

\section{Experiment and Results}
\subsection{Experimental Setup}
Our evaluation employs the Switch Transformer~\cite{fedus2022switch} as the MoE-based LLM backbone, which represents a classic scalable MoE design, utilizing a Top-1 routing that assigns tokens to one of 16 experts per layer.
To evaluate model performance, we report the test accuracy for the global model on widely used datasets, spanning from basic semantic classification (AGNews~\cite{zhang2015character}) to complex knowledge reasoning (MMLU~\cite{hendrycks2020measuring}).
The performance of \name is compared against several representative FL frameworks, including FedAvg~\cite{mcmahan17a}, PFL-MoE~\cite{guo2021pfl}, and FedMoE~\cite{mei2024fedmoe}.

The distributed system comprises a central server and 10 synchronized clients, with computations accelerated by NVIDIA GeForce RTX 4090 GPUs. For local fine-tuning, each client executes one epoch per communication round with a learning rate of $1\times10^{-4}$, with a total of 25 communication rounds.
The key hyper-parameters in \name is set to $\lambda = 0.5$.
To simulate real-world data heterogeneity in federated learning, we partition the datasets using a Dirichlet-based distribution scheme~\cite{guo2021pfl}. By setting the concentration parameter $\alpha = 0.1$, we induce significant label skew across the client-specific data distributions, providing a challenging non-IID setting for evaluating our proposed conflict-aware aggregation framework.

\begin{table*}[t]
\centering
\renewcommand{\arraystretch}{1.8}
  \begin{tabular}{c||ccc|ccc|ccc}
    \hline
    \hline
    \multicolumn{10}{c}{\textbf{AGNews}} \\
    \hline
    \multirow{2}{*}{Method} & \multicolumn{3}{c|}{\textbf{$\alpha=0.1$}} & \multicolumn{3}{c|}{\textbf{$\alpha=0.5$}} & \multicolumn{3}{c}{\textbf{$\alpha=1.0$}} \\
    \cline{2-10}
    & accuracy & precision & recall & accuracy & precision & recall & accuracy & precision & recall \\
    \hline
    \texttt{FedAvg} & 0.7740 & 0.7940 & 0.7568 & 0.8866 & 0.8921 & 0.8739 & 0.8978 & 0.9089 & 0.8907 \\
    \texttt{PFL-MoE} & 0.7994 & 0.8223 & 0.7911 & 0.9041 & 0.9128 & 0.9036 & 0.9109 & 0.9170 & 0.9106 \\
    \texttt{FedMoE} & 0.8137 & 0.8374 & 0.8056 & 0.9103 & 0.9192 & 0.9120 & 0.9164 & 0.9204 & 0.9155 \\
    \texttt{FC-MoE} & \textbf{0.8359} & \textbf{0.8617} & \textbf{0.8360} & \textbf{0.9194} & \textbf{0.9246} & \textbf{0.9194} & \textbf{0.9224} & \textbf{0.9230} & \textbf{0.9223} \\
    \hline
    \multicolumn{10}{c}{\textbf{MMLU}} \\
    \hline
    \multirow{2}{*}{Method} & \multicolumn{3}{c|}{\textbf{$\alpha=0.1$}} & \multicolumn{3}{c|}{\textbf{$\alpha=0.5$}} & \multicolumn{3}{c}{\textbf{$\alpha=1.0$}} \\
    \cline{2-10}
    & accuracy & precision & recall & accuracy & precision & recall & accuracy & precision & recall \\
    \hline
    \texttt{FedAvg} & 0.3008 & 0.3870 & 0.3002 & 0.3482 & 0.3854 & 0.3478 & 0.4012 & 0.4002 & 0.4010 \\
    \texttt{PFL-MoE} & 0.3325 & 0.3541 & 0.3301 & 0.3789 & 0.3947 & 0.3740 & 0.4305 & 0.4242 & 0.4275 \\
    \texttt{FedMoE} & 0.3506 & 0.3699 & 0.3500 & 0.3921 & 0.4063 & 0.3919 & 0.4463 & 0.4412 & 0.4460 \\
    \texttt{FC-MoE} & \textbf{0.3764} & \textbf{0.4017} & \textbf{0.3760} & \textbf{0.4169} & \textbf{0.4372} & \textbf{0.4168} & \textbf{0.4695} & \textbf{0.4650} & \textbf{0.4694} \\
    \hline
    \hline
  \end{tabular}
\caption{The test accuracy on the AGNews and MMLU datasets under non-IID data distributions.}
\label{tab:data_heterigeneity}
\vspace{-2ex}
\end{table*}

\subsection{Experimental Results and Analysis}
Table~\ref{tab:data_heterigeneity} summarizes the test accuracy across varying levels of data heterogeneity, where methods are evaluated under three non-IID configurations ($\alpha \in \{0.1, 0.5, 1.0\}$).
\name exhibits consistent superiority over all baselines on both the AGNews and MMLU benchmarks, achieving a test accuracy of \needrev{0.8359 on AGNews and 0.3764 on MMLU} in the most challenging non-IID scenario ($\alpha = 0.1$). Moreover, the performance gap between \name and existing methods widens as the data distribution becomes increasingly skewed. 
This observation confirms that our conflict-aware aggregation effectively mitigates the destructive interference caused by divergent local expert optimizations, leading to a more robust and generalized global model.

\begin{figure}[t]
\vspace{-3ex}
\centering
\subfloat[Convergence \label{fig:convergence}]{
\includegraphics[width=0.49\linewidth]{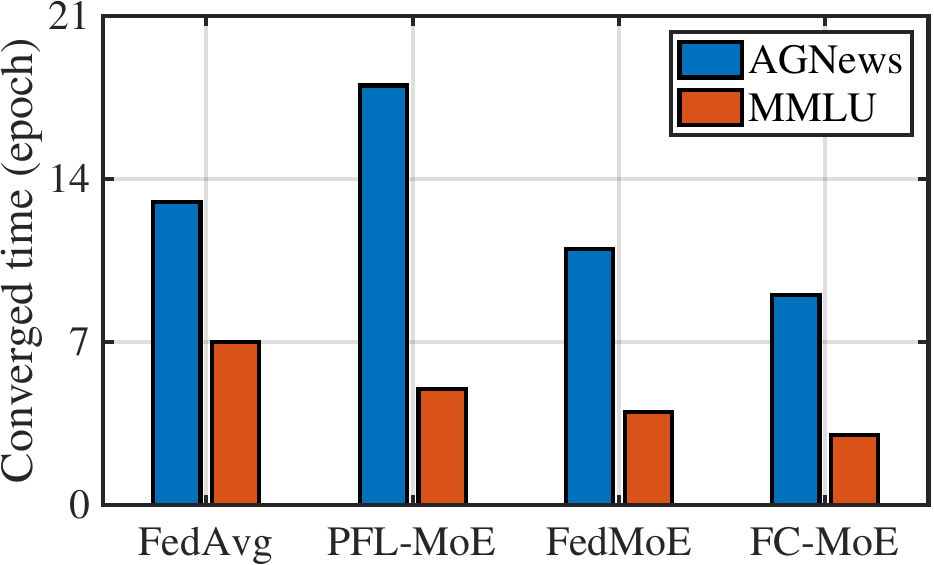}
}
\subfloat[Local accuracy \label{fig:local_accuracy}]{
\includegraphics[width=0.49\linewidth]{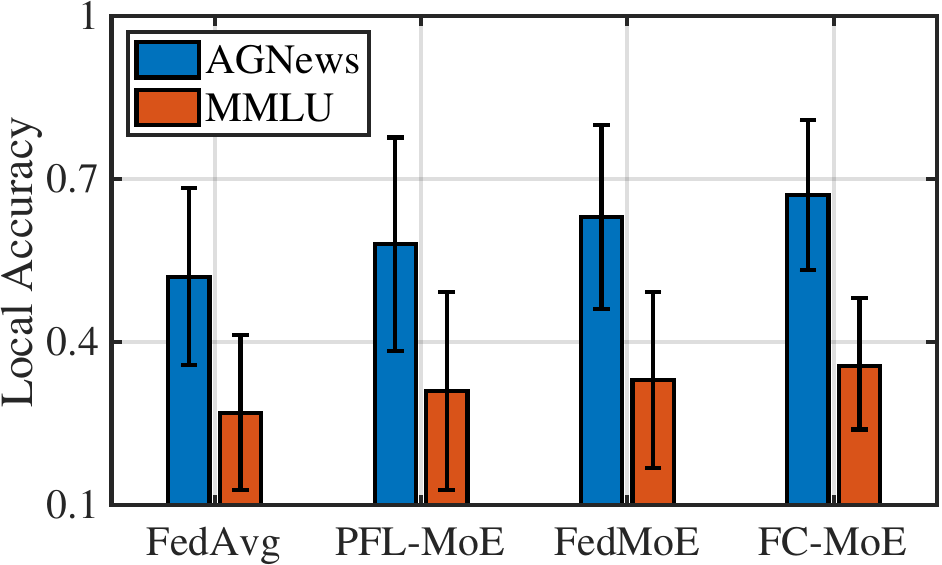}
}
\caption{The convergence and local accuracy across clients under non-IID data distributions with $\alpha=0.1$.}
\label{fig:local_performance}
\vspace{-1ex}
\end{figure}


Fig.~\ref{fig:convergence} illustrates the convergence rate for various benchmarks on the AGNews and MMLU datasets. Compared to other baselines, \name exhibits significantly faster convergence, reaching the target performance in fewer communication rounds. This acceleration is attributed to our gradient consensus projection. By ensuring aggregated updates align with the global consensus, \name maintains a more stable and efficient path toward the global optimum, effectively mitigating the optimization oscillations induced by gradient interference in heterogeneous environments.

To evaluate the model's capacity to preserve client-specific knowledge, we analyze the statistics of local accuracy on clients in Fig.~\ref{fig:local_accuracy}. The results reveal that \name not only achieves a higher mean accuracy but also exhibits substantially lower variance across clients, demonstrating superior local representational fidelity. This success is primarily due to the local knowledge retention strategy, which allows clients to re-anchor domain-specific residuals during local initialization, ensuring global alignment without compromising local expertise.


\begin{table}[t]
\centering
\renewcommand{\arraystretch}{1.8}
  \begin{tabular}{c||cc||cc}
    \hline
    \hline
    \multirow{2}{*}{Method} & \multicolumn{2}{c||}{\textbf{AGNews}} & \multicolumn{2}{c}{\textbf{MMLU}} \\
    \cline{2-5}
    & test acc & local acc & test acc & local acc \\
    \hline
    \texttt{a)} & 0.7849 & 0.5574 & 0.3306 & 0.2992 \\
    \texttt{b)} & 0.8068 & 0.6108 & 0.3478 & 0.3225 \\
    \texttt{c)} & 0.8264 & 0.6459 & 0.3616 & 0.3428 \\
    \texttt{FC-MoE} & \textbf{0.8359} & \textbf{0.6732} & \textbf{0.3764} & \textbf{0.3559} \\
    \hline
    \hline
  \end{tabular}
\caption{The test and local accuracy for \name under non-IID data distributions with $\alpha=0.1$.}
\label{tab:ablation}
\vspace{-3ex}
\end{table}

\begin{figure}[t]
\centering
\subfloat[AGNews]{
\includegraphics[width=0.49\linewidth]{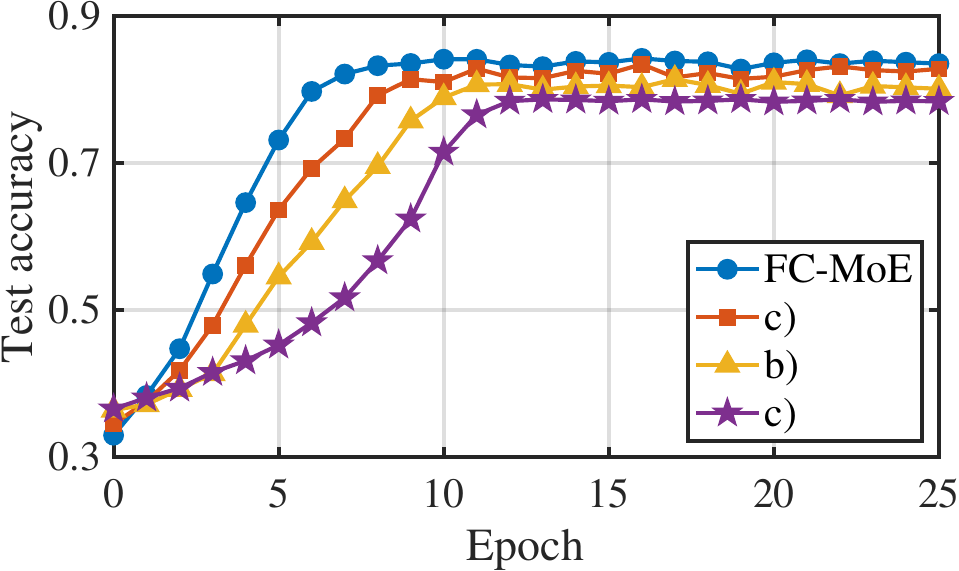}
}
\subfloat[MMLU]{
\includegraphics[width=0.49\linewidth]{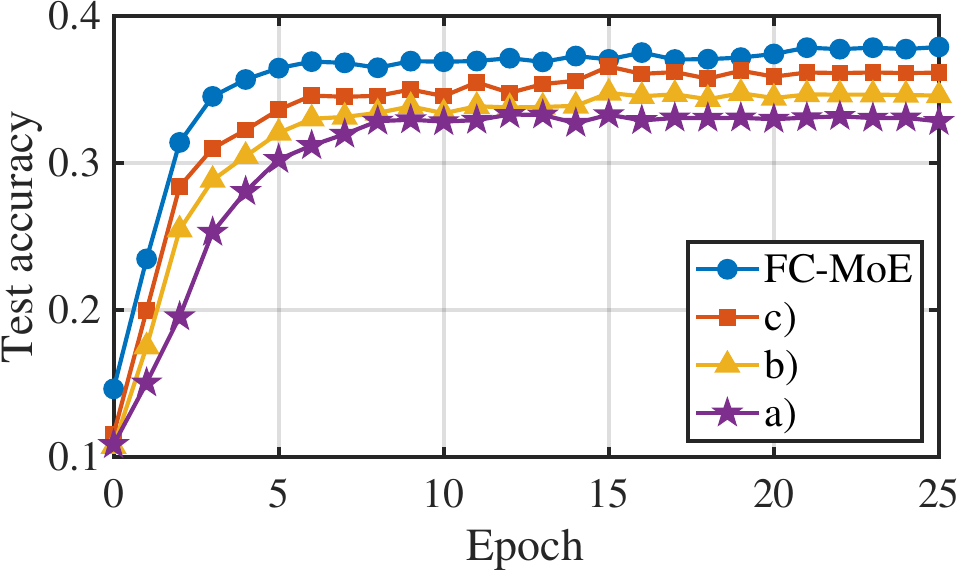}
}
\vspace{-1ex}
\caption{The test accuracy of \name and its variants on AGNews and MMLU datasets.}
\label{fig:ablation}
\vspace{-1ex}
\end{figure}

\subsection{Ablation Study}
We conduct the ablation study to quantify the individual contributions of our designed components, with results visualized in Table~\ref{tab:ablation} and Fig.~\ref{fig:ablation}, where \name is compared against three degraded variants: a) without gradient consensus projection, b) without local knowledge retention, and c) without importance-aware expert weighting. The results demonstrate that the removal of any single component leads to noticeable performance drops and increased oscillation in the test accuracy. The comparison confirms that the gradient consensus projection is the primary driver of optimization stability, while the importance-aware expert weighting ensures the consensus direction is driven by more reliable local updates. Further supported by local knowledge retention, \name effectively balances global convergence and local specialization in heterogeneous environments.


\section{Conclusion}
In this paper, we have proposed a novel conflict-aware aggregation framework, named \name, designed to tackle the inherent expert conflicts in federated MoE-based LLM fine-tuning under non-IID data distributions. 
\name explicitly quantifies the reliability of local updates through an importance-aware weighting scheme and geometrically aligns conflicting experts via a gradient consensus projection. Furthermore, a local knowledge retention mechanism is introduced to strategically anchor the filtered domain-specific residuals, preventing the degradation of client personalization.
Extensive experiments across diverse benchmarks demonstrate the superiority of \name, which accelerates global model convergence and enhances generalization without sacrificing local expert specialization.



\bibliographystyle{IEEEtran}
\bibliography{reference}

\end{document}